%% file: main.tex
\documentclass[conference]{IEEEtran}
\IEEEoverridecommandlockouts
\usepackage{cite}
\usepackage{amsmath,amssymb,amsfonts,bm}
\usepackage{graphicx}
\usepackage{textcomp}
\usepackage{xcolor}
\usepackage{float}
\usepackage{algorithm}
\usepackage{algorithmicx}
\usepackage{algpseudocode}
\usepackage{makecell}
\usepackage{verbatim}

\usepackage{caption}
\usepackage{subfigure}
\usepackage{mdwlist}
\usepackage{enumitem}
\usepackage{xcolor}
\usepackage[utf8]{inputenc}
\usepackage[english]{babel}

\def\BibTeX{{\rm B\kern-.05em{\sc i\kern-.025em b}\kern-.08em
    T\kern-.1667em\lower.7ex\hbox{E}\kern-.125emX}}
    
\begin{document}

\title{Fast Approximate Spectral Normalization for Robust Deep Neural Networks
}

% \author{\IEEEauthorblockN{Zhixin Pan}
% \IEEEauthorblockA{\textit{CISE,UFL}\\
% \textit{Gainesville, FL}\\
% %Gainesville, US \\
% panzhixin@ufl.edu}
% \and
% \IEEEauthorblockN{Ziyu Shu}
% \IEEEauthorblockA{\textit{CISE,UFL}\\
% \textit{Gainesville, FL}\\
% ziyushu@ufl.edu}
% \and
% \and

\author{\IEEEauthorblockN{Zhixin Pan and Prabhat Mishra}
\IEEEauthorblockA{Department of Computer \& Information Science \& Engineering\\
University of Florida, Gainesville, Florida, USA}}

\maketitle

\begin{abstract}
Deep neural networks (DNNs) play an important role in machine learning due to its outstanding performance compared to other alternatives. However, DNNs are not suitable for safety-critical applications since DNNs can be easily fooled by well-crafted adversarial examples. One promising strategy to counter adversarial attacks is to utilize spectral normalization, which  ensures that the trained model has low sensitivity towards the disturbance of input samples. Unfortunately, this strategy requires exact computation of spectral norm, which is computation  intensive and impractical for large-scale networks. In this paper, we introduce an approximate algorithm for spectral normalization based on Fourier transform and layer separation. The primary contribution of our work is to effectively combine the sparsity of weight matrix and decomposability of convolution layers. Extensive experimental evaluation demonstrates that our framework is able to significantly improve both time efficiency (up to 60\%) and model robustness (61\% on average) compared with the state-of-the-art spectral normalization. 
% should come with some statistics
\end{abstract}


\input{sections/intro.tex}
\input{sections/bgd}

\input{sections/accelerate.tex}
\input{sections/exp}
\input{sections/conclusion.tex}

\bibliographystyle{IEEEtran}
\bibliography{zhixin.bib}

\end{document}

%% file: sections/intro.tex
\section{Introduction}
% Introduce to DNN especially CNN, adversarial attack, show some scenario where attack take place. Then comes to Spectral Norm Regulization

% Relate part should be involved, such as FSGM and DeepFool, and mention the paper dealing with spectral Norm , show their disadvantages.

\begin{figure}[htbp]
\centering
\includegraphics[scale =0.32]{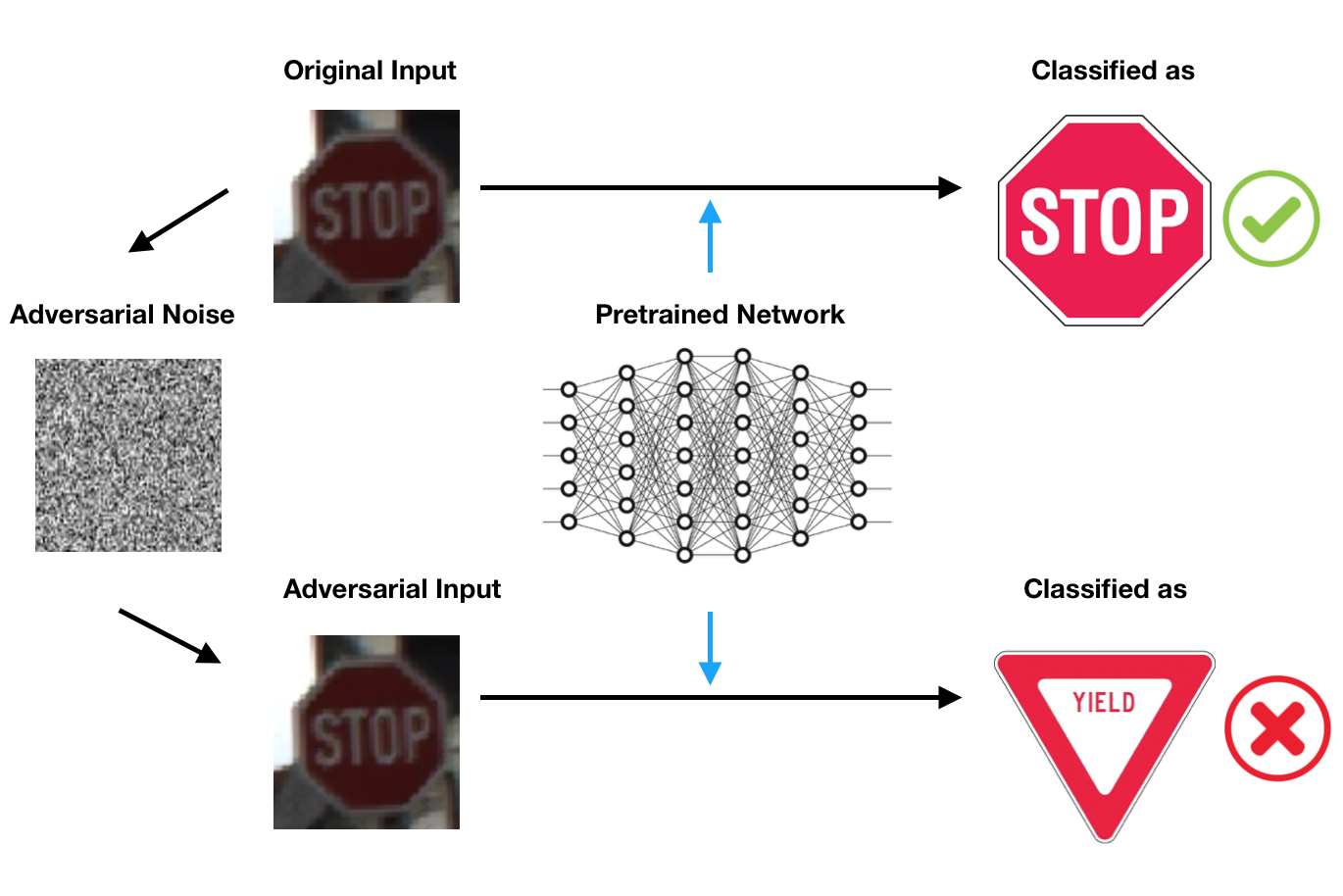}
\vspace{-0.15in}
\caption{Example adversarial attack on autonomous driving}
\label{fig:label1}
\vspace{-0.2in}
\end{figure}

% This part is too short, need to be refined 
Deep neural networks (DNNs) are are widely used in machine learning field. They can express or simulate a wide variety of intrinsic functionalities including classification, regression, reconstruction, etc. The flexibility of DNNs also enables their different variations to be successfully employed in diverse applications. 
% For example, \textit{convolution neural networks}(CNN) is a class of DNNs with impressive capabilities on image pattern recognition and has been widely used in Computer Vision algorithms including visual object
% classification. \textit{Recurrent neural network}(RNN) is capable of handling long-term time-based input sequences, and therefore became the state-of-the-art method for \textit{natural language processing}(NLP) problems. \textit{Generative adversarial networks}(GAN) proposed by Ian Goodfellow and his colleagues has already gained significant progress in simulation and reconstruction works such as Deepface, gravitational lensing simulation. 
% \begin{figure}[htbp]
% \centering
% \includegraphics[scale =0.3]{sections/figs/fig1_apps.png}
% \caption{Applications of Deep Neural Networks}
% \label{fig:label0}
% \end{figure}
The adversarial attack proposed by Szegedy et al.\cite{Sze2014} revealed the vulnerability of most existing neural networks against adversarial examples. 
%Adversarial samples are a type of sample that is maliciously designed to attack a machine learning model. 
The difference between adversarial samples and the original sample can hardly be distinguished by naked eyes, but will lead the model to make an incorrect prediction with high confidence.
As shown in Figure~\ref{fig:label1}, a human-invisible noise was added to input traffic sign image. While a pre-trained network can successfully recognize the original input as a stop sign, the same network will incorrectly classify it as a yield sign if the input is perturbed with well-crafted noise. There are many such real-world examples of malicious (adversarial) attacks. In order to design robust DNNs, it is critical defend against adversarial attacks.

%\vspace{-0.15in}

% Regardless of all existed exhilarating contributions, recent study revealed neural networks' serious vulnerability against \textit{adversarial attacks}. That is, a tiny perturbation on the original input, which is usually imperceptible to human, may cause a significant difference in the output and fools the model easily.

While there are many promising defense strategies, most of them are designed for specific attack algorithms, which severely restricts their applicability. Spectral normalization \cite{SNCNN} has  gained significant attention because it is algorithm-agnostic and can reduce DNNs' sensitivity to input perturbation. A major challenge with this method is that it provides a trade-off between computation time  and numerical accuracy for computing the spectral norm of DNNs' weight matrix. Specifically, it applies one iteration of power method for each individual layer, which achieves poor accuracy in most cases. Moreover, it introduces high computation overhead when dealing with  large convolution kernels. In reality, power iteration may not numerically converge to the desired result in some specific scenarios. 

In this paper, we propose a fast and dependable acceleration approach for spectral normalization. We have evaluated the effectiveness of our proposed approach using several benchmark deep neural networks along with datasets. Specifically, this paper makes the following three important contributions.

\begin{enumerate}
    \item To the best of our knowledge, we are the first to utilize spatial separation of convolution layers for regularized  training.
    \item We develop a fast approximate spectral norm computing scheme for convolution layers, which is based on Fourier transform and can be easily extended to fully-connected layers.
    \item We demonstrate the feasibility of our approach with experiments on a variety of classic network structures and dataset. Experimental results demonstrate the improvement of DNNs' robustness against adversarial attack through extremely low attack success rate for bounded attack and high distortion for unbounded attack. 
\end{enumerate}

The rest of this paper is organized as follows: We firstly provide background on adversarial attack and spectral norm regularization in Section 2. Section 3 describes our proposed method where we will demonstrate detailed algorithm workflow. Experimental results are provided in Section 4, and we conclude the paper in Section 5.

%% file: sections/bgd.tex
\section{Background and Related Work}
\label{bgd}
In this section, we provide relevant background before describing related efforts.
First, we 
%provide preliminary knowledge about adversarial attacks. Then we 
describe three types of layers in a DNN. Next, we define a given model's stability using linear algebra concepts. Finally, we discuss related efforts to motivate the need for our proposed approach. 
%, we will illustrate how neural networks' forward pass is closely related to matrix multiplication, by which we can 

% \subsection{Adversarial Attack}
% Even though Deep learning has earned much reputation as a powerful learning approach applied to a broad spectrum of tasks, the adversarial attack proposed by Szegedy et al.\cite{Sze2014} revealed the vulnerability of most existing neural networks against adversarial examples. Adversarial samples are a type of sample that is maliciously designed to attack a machine learning model. The difference between them and the original sample can hardly be distinguished by naked eyes, but will lead the model to give a false output with high confidence.

% As shown in Figure~\ref{fig:label1}, a human-invisible noise was added to input traffic sign image. While a pre-trained network can successfully recognize the original input as a stop sign, the same network will incorrectly classify it as a yield sign if the input is perturbed with well-crafted noise. There are many such real-world examples of malicious (adversarial) attacks. The existence of adversarial samples make the application of deep learning in fields of security seriously threatened, and how to effectively defend them is an important research topic.

% \vspace{-0.15in}
% \begin{figure}[htbp]
% \centering
% \includegraphics[scale =0.32]{sections/figs/Figure2_new.png}
% \vspace{-0.15in}
% \caption{Example of adversarial attack on Autonomous Driving System}
% \label{fig:label1}
% \vspace{-0.1in}
% \end{figure}

\subsection{Three Types of Layers in DNNs}
DNN consists of the following three types of layers: fully connected layer, convolution layer and activation functions.

\begin{figure}[htbp]
\vspace{-0.15in}
\centering
\includegraphics[scale =0.5]{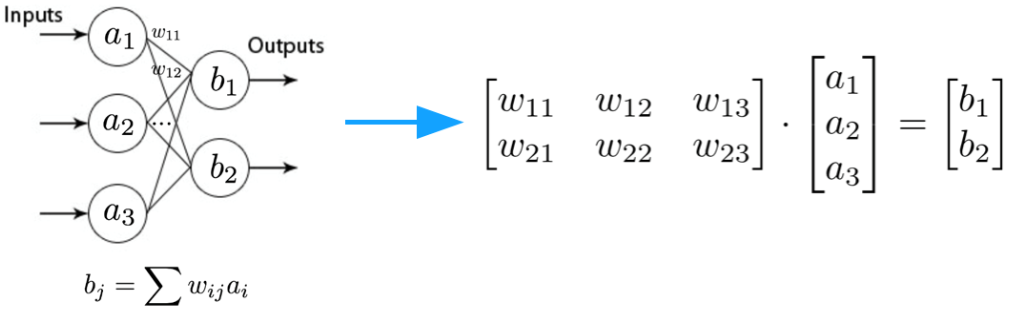}
\vspace{-0.25in}
\caption{Fully connected layer in DNNs}
\vspace{-0.2in}
\label{fig:label3}
\end{figure}

\vspace{0.1in}
\noindent {\it Linear Layer: }
In this layer, each output node is nothing but a weighted sum of inputs as shown in Figure~\ref{fig:label3}. If we consider all equations and view it as a linear system, it can be represented in the form of matrix multiplication.

\begin{figure}[htbp]
\centering
\vspace{-0.1in}
\includegraphics[scale =0.265]{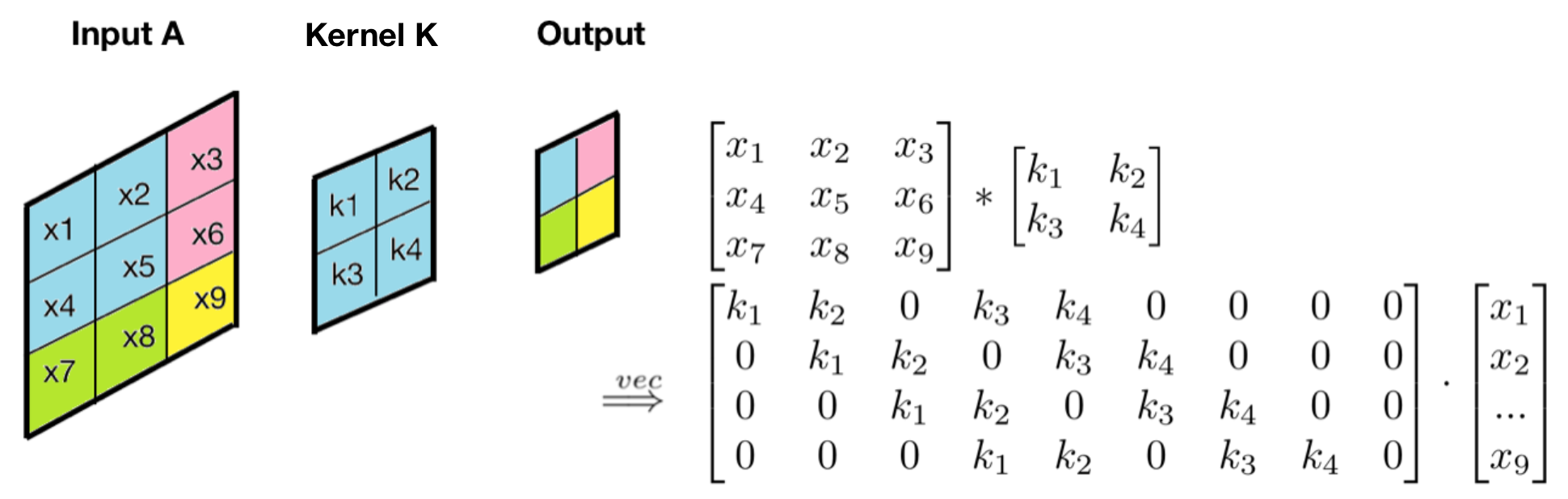}
%\vspace{-0.25in}
\caption{Convolution layer in DNNs}
\vspace{-0.1in}
\label{fig:label4}
\end{figure}

\vspace{0.1in}
\noindent {\it Convolution Layer:}
Convolution layers are widely applied in computer vision tasks, where a convolution kernel $K$ will slide along the surface of input feature map, and each weighted sum is stored into an entry of the output feature map. We can represent it as matrix multiplication between the input vector and a doubly block circulant matrix (which is called the convolution matrix of $K$). 
Figure~\ref{fig:label4} shows a trivial example to illustrate this idea. 
% \[
% \begin{split}
% & \begin{bmatrix}
% x_1 & x_2 & x_3\\
% x_4 & x_5 & x_6\\
% x_7 & x_8 & x_9\\
% \end{bmatrix} *
% \begin{bmatrix}
% k_1 & k_2\\
% k_3 & k_4
% \end{bmatrix}\\
% \overset{vec}{\Longrightarrow{}}&
% \begin{bmatrix}
% k_1 & k_2 & 0 & k_3 & k_4 & 0 & 0 & 0 & 0\\
% 0 & k_1 & k_2 & 0 & k_3 & k_4 & 0 & 0 & 0\\
% 0 & 0 & k_1 & k_2 & 0 & k_3 & k_4 & 0 & 0\\
% 0 & 0 & 0 & k_1 & k_2 & 0 & k_3 & k_4 & 0
% \end{bmatrix}
% \cdot
% \begin{bmatrix}
% x_1 \\ x_2 \\ ... \\ x_9
% \end{bmatrix}\\
% = & \begin{bmatrix}
% k_1x_1+k_2x_2+k_3x_4+k_4x_5 \\
% k_1x_2+k_2x_3+k_3x_5+k_4x_6 \\
% k_1x_4+k_2x_5+k_3x_7+k_4x_8 \\
% k_1x_5+k_2x_6+k_3x_8+k_4x_9
% \end{bmatrix}
% \end{split}
% \]

%\vspace{0.1in}
\noindent {\it Activation Functions: }
Activation functions are inserted between consecutive network layers to induce nonlinearity to allow DNNs complete nontrivial tasks. They are usually selected as piecewise linear functions, such as ReLU, maxout, and maxpooling. Since adversarial examples always reside in the close neighborhood of a given input, activation function can be approximately considered as a linear function in this case, which is compatible with matrix representation. 

\subsection{Stability of DNNs}
Since we have already shown the close relationship between DNNs' layer and matrix multiplication, we can interpret adversarial attacks into the language of linear algebra. Assume that our model's interpreted matrix (we will call this \textit{weight matrix} in the rest of this paper) is $W \in \mathbb{R}^{m \times n}$, and there is a well-crafted noise $\xi \in \mathbb{R}^{n}$ added to our input vector $x \in \mathbb{R}^{n}$, then we measure change of output by the following relative error $\epsilon \in \mathbb{R}$:
\[
\epsilon = \frac{||W \cdot (x + \xi) - W \cdot x||_2}{||\xi||_2} = \frac{||W \cdot \xi||_2}{||\xi||_2}
\]
Notice the spectral norm $\sigma(W)$ of matrix $W$ is defined as:
\[
\sigma(W)  \triangleq \max \limits_{\xi \neq \textbf{0}}\frac{||W\cdot \xi||_2}{||\xi||_2}
\]
Therefore, spectral norm of matrix $W$ gives a tight upperbound of given layer's stability, and a small value of $\sigma(W)$ indicates this model's insensitivity to the perturbation of input $x$. 

\subsection{Related Work}
\label{relw}
Based on matrix representation of DNN' forward pass, recent research efforts have developed several countermeasures, but many of them failed to protect against strong attacks~\cite{CarliniW16a, DeepFool, DecisionTreeAtk, FSGM, IterativeAtk}. \textit{Spectral Normalization} (SN)~\cite{SNCNN} is one of the most promising defense strategies because it is algorithm-agnostic approach and is able to enhance DNNs' robustness against adversarial attacks.
During model training, the key idea of spectral normalization is to append the spectral norm of each layer to loss function as penalty term, so that it can reduce DNNs' sensitivity to input perturbation by minimizing every layer's spectral norm. The loss function is:
\[ J = \frac{1}{N}\sum \limits_{i=1}^{N} L(f(x_i), y_i) + \frac{\lambda}{2}\sum\limits_{k=1}^K\sigma(W^k)^2 \]
Here, $x_i$ is a training sample, the label of $x_i$ is denoted as $y_i$,  the total number of training samples in a batch is denoted as $N$, $L$ is the dissimilarity measurement and is frequently selected to be cross entropy or squared $l_2$ distance, $\sigma(W^k)$ is the spectral norm of $k$-th layer, $K$ is the total number of layers, and $\lambda \in \mathbb{R}^+$ is a regularization factor.

The state-of-the-art method in~\cite{SNCNN} applied power iteration to approximate $\sigma(W^k)$, then perform \textit{stochastic gradient descent} (SGD) to train the DNN, as shown in Algorithm~\ref{Alg1}.

\begin{algorithm}
    \caption{State-of-the-art: Spectral Normalization (SN)}
    \begin{algorithmic}[1] % show lines 
        \For {each iteration of SGD}
            \For {$k = 1$ to $K$}
                \For {a sufficient number of iterations}
                    \State Apply power iteration to compute $\sigma(W^k)$
                    \State Add $\sigma(W^k)$ to loss function
                \EndFor
            \EndFor
            \State Compute gradient of modified loss function
            \State Update parameters by back propagation
        \EndFor
    \end{algorithmic}
    \label{Alg1}
\end{algorithm}

This spectral normalization (SN) algorithm suffers from high computation complexity. Consider a convolution layer with $a$ input channels, $b$ output channels, and a convolution kernel in a size of $w \times h$. The corresponding weight matrix will be at least in a size of $b \times awh$. The exact computation of its spectral norm requires SVD decomposition, whose complexity is $\bm{O}(min(m^2n, n^2m))$ for a $m \times n$ matrix, Therefore the time complexity in our task is $\bm{O}(min(a^2bw^2h^2, ab^2wh))$, which is infeasible to apply during network training. In order to make it run in a reasonable time, the authors in~\cite{SNCNN} use only a few iterations of power iteration method to approximate the spectral norms which leads to very coarse approximation. As a result, it significantly compromises its robustness goal. Our proposed method provides a fast approximate algorithm for spectral normalization through synergistic integration of layer separation and Fourier transform as described in the next section.

%\color{blue}

%\color{black}

% \begin{figure}[htbp]
% \centering
% \vspace{-0.15in}
% \includegraphics[scale =0.38]{sections/figs/visualization.png}
% \vspace{-0.2in}
% \caption{Comparison between existing and proposed method.}
% \label{fig:concept}
% \vspace{-0.1in}
% \end{figure}

%Figure~\ref{fig:concept} compares state-of-the-art spectral normalization (left) with our proposed framework (right). 

%% file: sections/accelerate.tex
\section{Fast Approximate Spectral Normalization}

\begin{figure}[htbp]
\centering
\includegraphics[scale =0.26]{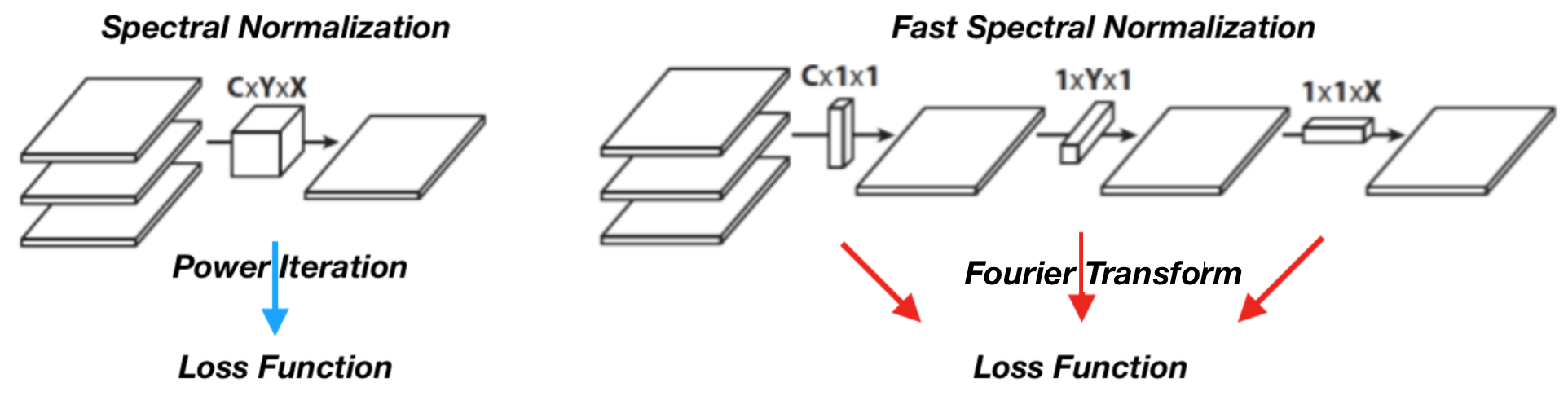}
\caption{Comparison of our proposed framework with the state-of-the-art. We utilize layer separation to decompose large and multi-dimension layers into 1-D layers, then apply Fourier transform on each of the kernels.}
%\vspace{-0.1in}
\label{fig:label5}
\end{figure}

%In this section, we explain our approach to refine the process for the fast spectral normalization. 
Figure~\ref{fig:label5} shows an overview of our proposed framework for \textit{fast spectral normalization} (FSN). The major steps of our framework is outlined in Algorithm~\ref{Alg3}. Our proposed framework consists of two strategies to accelerate training process: \textit{layer separation} and \textit{Fourier transform}. The remainder of this section describes these strategies in detail.
\vspace{-0.1in}

\subsection{Layer Separation}
\label{sec:ls}
The reason for computing spectral norm being so expensive is the size of weight matrix. Here we apply \textit{layer separation} to reduce the time complexity for this task. First, a 2-D filter kernel $K$ is said to be $separable$ if it can be expressed as the outer product of one row vector $\bm{r}$ and one column vector $\bm{c}$. But vectors are special cases of matrix, and their outer product is equivalent to their convolution. Then due to the associativity of convolution:
\[ 
\begin{split}
&K = \bm{r}\times \bm{c} = \bm{r}*\bm{c}\\
\Rightarrow{}& A * K = A * (\bm{r} * \bm{c}) = (A * \bm{r}) * \bm{c} 
\end{split}
\]

A famous example will be the Sobel kernel, which is widely used to approximate the gradient value of image brightness function:
\[
\begin{bmatrix}
-1 & 0 & 1\\
-2 & 0 & 2\\
-1 & 0 & 1
\end{bmatrix}
= 
\begin{bmatrix}
1\\
2\\
1
\end{bmatrix}
\times
\begin{bmatrix}
-1 & 0 & 1\\
\end{bmatrix}
\]

Once the given convolution layer is separable ($\operatorname{rank}(W) = 1$),  we can always decompose it into 2 1-D layers but achieve the same effect. This time we have 2 consecutive convolution layers, and their kernel size will be $w\times1$ and $1\times h$. We can compute each layer's spectral norm separately and then append them to the loss function together.

What about inseparable layers? In this case, our method utilize the SVD decomposition of kernel filter to transfer it into its low-rank approximation. Here, we are performing SVD decomposition on kernel matrix instead of the doubly block circulant matrix. Kernel matrix encountered in modern CNNs are usually in the form of $3 \times 3$ or $5 \times 5$, time cost for SVD decomposition on these tiny matrices is trivial. Also, applying low-rank approximation of kernel filter is reasonable when it comes to real situations. In CNN applications, especially in computer vision areas, many frequently used feature extraction kernels like edge detection filters are trained to have one dominant eigen direction, where the largest singular value is far larger than the inferior ones, one extreme example is the Sobel kernel mentioned above. Under this circumstance, the deviation induced by low-rank approximation can be neglected. 

\subsection{Fourier Transform}
\label{sec:Fourier}
It is a well known fact that the spectral norm of a 2D convolution kernel $K$ is exactly the largest eigenvalue of $A^{\top}A$, where $A$ is the corresponding convolution matrix of $K$. In the baseline method illustrated in Algorithm~\ref{Alg1}, power iteration method was applied. But power method, as a fundamental numerical method for computing eigenvalues, usually takes too many iterations until acceptable accuracy is obtained. Also, it assumes the weight matrix has an eigenvalue that is strictly greater in magnitude than the others, and the initial random vector should contain an nonzero component in the direction of a dominant eigenvector. If the above assumptions fails, the power method may not converge!
Instead, we apply the following theorem \cite{sedghi2018singular} to calculate the spectral norm of a convolution kernel fast.
% \cite{sedghi2018singular}
%\begin{theorem}

\textbf{Theorem 1:} For any convolution matrix formed by kernel $K$, the eigenvalues of the convolution matrix are the entries of 2D Fourier transform of $K$, and its singular values are their magnitudes.
%\end{theorem}

\textbf{Proof:} Assume $K$ be a kernel matrix, and $A$ is the convolution matrix of $K$ (as described in Figure~\ref{fig:label4}). Now the task is to determine the singular values of a doubly block circulant matrix $A$. First, we define
\[
Q \triangleq \frac{1}{n} (F \otimes F)
\]
where $F$ is the Fourier matrix. To complete the proof, we use the following lemma.

\textbf{Lemma 1:}
For any doubly block circulant matrix $A$, the eigenvectors of $A$ are the columns of $Q$, and $Q$ is unitary. (J.Toriwaki(1989) in \cite{Jain})

Based on the above lemma, we know $A$ can always be decomposed into its eigenvalue decomposition $A = Q*DQ$ where D is a diagonal matrix. Now we show that $A$ is normal: 
\[
\begin{split}
 &AA^\top = AA^* = Q^*DQQD^*Q \\
= & Q^*DD^*Q = Q^* D^* DQ = Q^*D^*Q Q^*DQ\\
= & A^*A = A^\top A    
\end{split}
\]

Since the singular values of any normal matrix are the magnitudes of its eigenvalues (Johnson (2012), page 158~\cite{Horn}) and we have shown that $A$ is normal, by applying the Lemma~1 we complete the proof. $\blacksquare$ 

Therefore, we can calculate the spectral norm of target convolution kernel by Fourier transform, and this approach is depicted in Algorithm~\ref{Alg2}.

\begin{algorithm}[ht]
\caption{Algorithm for computing spectral norms}
\hspace*{0.02in} {\bf Input:} 
\hspace*{0.02in} convolution kernel $K$\\
\hspace*{0.02in} {\bf Output:} $\sigma$, the spectral norm respected to $K$.
\begin{algorithmic}[1]
            \State $k \leftarrow$ 2D Fourier transform of $K$
            \State $k' \leftarrow$ set all the entries of $k$ to zero except the one with maximum absolute value 
            \State $K' \leftarrow$ 2D inverse Fourier transform of $k'$
            \State $\sigma \leftarrow$ largest entry inside $K'$.
\end{algorithmic}
\label{Alg2}
\end{algorithm}

\subsection{Activation Functions}
By exploiting layer separation and Fourier transform, the computation of spectral norm for linear and convolution layers can be solved efficiently as shown in Algorithm~\ref{Alg2}. When it comes to activation layers, there is no inherent way to represent activation layers in matrix format. Activation functions are deployed in neural networks to induce nonlinearity, therefore it is impossible to view an activation function as a linear transformation. 

To perform our method with activation functions, instead of parsing it into matrix and compute spectral norm, we take their \textit{Lipschitz constants} into consideration. A function $f$ defined on $\textbf{X}$ is said to be $K-$Lipschitz if 
\[
\forall x_1,x_2 \in \textbf{X},\,|f(x_1)-f(x_2)| \leq K|x_1 - x_2| 
\]
The smallest possible constant $K$ for $f$ is also called \textit{Lipschitz norm}, which reflects how expansive function $f$ is. The Lipschitz norm upper bounds the relationship between input perturbation and output variation for a given distance. Notice the similarity between the definition of spectral norm and Lipschitz norm. Actually, spectral norm of matrix $W$ is essentially the Lipschitz norm of function $f$ if $f(x) = Wx$.
% \[
% \begin{split}
% \sigma(W) = & \max \limits_{\xi \neq \textbf{0}}\frac{||W\cdot \xi||_2}{||\xi||_2} \\
% = & \sup \limits_{x_1 \neq x_2} \frac{||W\cdot (x_1 - x_2)||_2}{||x_1 - x_2||_2} \\
% = & \sup \limits_{x_1 \neq x_2} \frac{|f(x_1) - f(x_2)|}{|x_1 - x_2|}
% \end{split}
% \]

It has been proven that most activation functions such as ReLU, Leaky ReLU, SoftPlus, Tanh, Sigmoid, ArcTan or Softsign, as well as max-pooling, are \textit{short maps}~\cite{Lips}, i.e, they have a Lipschitz constant equal to 1. We refer the reader to ~\cite{DeepLearning} for detailed proof on this subject. As a result, the output variation induced by activation layers is already restricted by a constant number. Therefore, there is no need to spare extra effort to append regularization terms for activation functions to loss function, since it will not get changed during training. Moreover, the constant number is 1, which means activation functions will neither expand or contract the variation of layer outputs. In other words, we can omit activation layers when evaluating the stability of DNNs.

%\subsection{Summary of Algorithm}
\subsection{Complexity Analysis}
The complete framework for approximate spectral normalization is depicted in Algorithm~\ref{Alg3}. For our target problem, based on the discussion in Section~\ref{relw}, we decompose the $b\times awh$ weight matrix into 2 matrices, $b\times aw$ and $b\times ah$ for each. Then we apply Fourier transform on each separated matrix. In our experiment \textit{fast Fourier transform} (FFT) is applied with $\bm{O}(mnlog(mn))$ time complexity for $m \times n$ matrix. So the proposed method's complexity is $O(abw\log(abw) + abh\log(abh)))$.

\begin{algorithm}
    \caption{Proposed: Fast Spectral Normalization (FSN)}
    \begin{algorithmic}[1] % show lines 
        \For {each iteration of SGD}
            \State Compute the gradient of general loss function as usual
            \For {$k = 1$ to $K$}
            \If{Convolution Layer}
            \State Perform layer separation
            \State Form the corresponding convolution matrix
            \EndIf
            \If{Linear Layer}
            \State Form the corresponding weight matrix
            \EndIf
            \State Perform \textbf{Algorithm2} to compute $\sigma(W^k)$
            \State Add $\sigma(W^k)$ to loss function
            \EndFor
            \State Update parameters using modified gradient
        \EndFor
    \end{algorithmic}
    \label{Alg3}
\end{algorithm}

%% file: sections/exp.tex
\section{Experiments}
\label{exp}
We evaluated the effectiveness of our optimized spectral norm regularization framework to confirm its time-efficiency and ability of enhancing DNNs' robustness.

\subsection{Experiment setup}
Experiments were conducted on a host machine with Intel i7 3.70GHz CPU, 32 GB RAM and RTX 2080 256-bit GPU. We developed code using Python for model training. We used PyTorch as the machine learning library. For adversarial attack algorithms, we utilized the \textit{Adversarial Robustness 360 Toolbox} (ART)~\cite{art2018}. Based on this environment, we considered the following two settings for our experimental evaluation:

% We implement our RFB Net detector based on the framework of Pytorch
% , utilizing several parts of open source infrastructures provided by the ssd.pytorch
% repository2
% . Our training strategies mostly follow SSD, including data augmentation, hard negative mining, scale and aspect ratios for default boxes, and loss
% functions (e.g., smooth L1 loss for localization and softmax loss for classification),
% while we slightly change our learning rate scheduling for better accommodation
% of RFB. More details are given in the following section of experiments. All new
% conv-layers are initialized with the MSRA method
\begin{enumerate}
    \item The VGG16 classifier for \textit{German Traffic Sign Benchmark} (GTSB) dataset~\cite{GTSB}.
    \item The Lenet-5 network for \textit{Udacity Self-Driving} (USD) dataset~\cite{USD}.
\end{enumerate}

For each setting we compare the following three approaches:
\begin{enumerate}
\item \textbf{Normal}: Ordinary training without regularization, which is considered as control group.
\item \textbf{SN}: State-of-the-art approach (Algorithm 1) with spectral normalization~\cite{SNCNN}.
\item \textbf{FSN}: Our proposed approach (Algorithm 2) with fast and robust spectral normalization.
\end{enumerate}

As for result evaluation, we first evaluated the functionality of them by reporting their accuracy and training time. Next, both bounded and unbounded adversarial attacks were deployed to test their robustness. Finally, we tuned the regularizer factor $\lambda$ for FSN to demonstrate its stability across a wide variety of different $\lambda$ values.

%\vspace{-0.2in}
\subsection{Case Study: German Traffic Sign Benchmark (GTSB)}
% Focus should be on GTSB instead of VGG, just mention kernels.
We train VGG16 on GTSB dataset. GTSB is a large and reliable benchmark which contains more than 50000 traffic sign image samples over 43 classes. VGG16~\cite{Alex} contains 16 hidden layers (13 convolution layers and 3 fully connected layers). One advantage of VGG16 is the replacement of larger convolution kernels ($11\times11, 7\times7, 5\times5$) with consecutive $3\times3$ convolution kernels, which makes it coincidentally suitable for our framework since small size of convolution kernel implies less time cost for low-rank approximation.
We choose a mini-batch size of 64. For each setting, we train the network with 200 epochs. For every 10 epochs, we randomly shuffle and split $80\%$ as training set and $20\%$ as test set, and then report the training time along with test performance.

\vspace{-0.1in}
\begin{table}[htbp]
    \caption{Training time and test accuracy of VGG16}
    %\vspace{20pt}
    \centering
    \begin{tabular}{c|cc}
        \hline
        \thead[l]{Methods} & \thead[l]{Time(s/epoch)}& \thead[l]{Best Test Accuracy(\%)} \\
        \hline
        Normal  & 14.5053   & 83.45 \\
        SN  & 29.1126   & 91.72 \\
        FSN & 18.1345   & 95.46 \\
        \hline       
    \end{tabular}
    \label{Tb1}
\end{table}
\vspace{-0.05in}

\begin{figure}[htbp]
\centering
\includegraphics[scale =0.17]{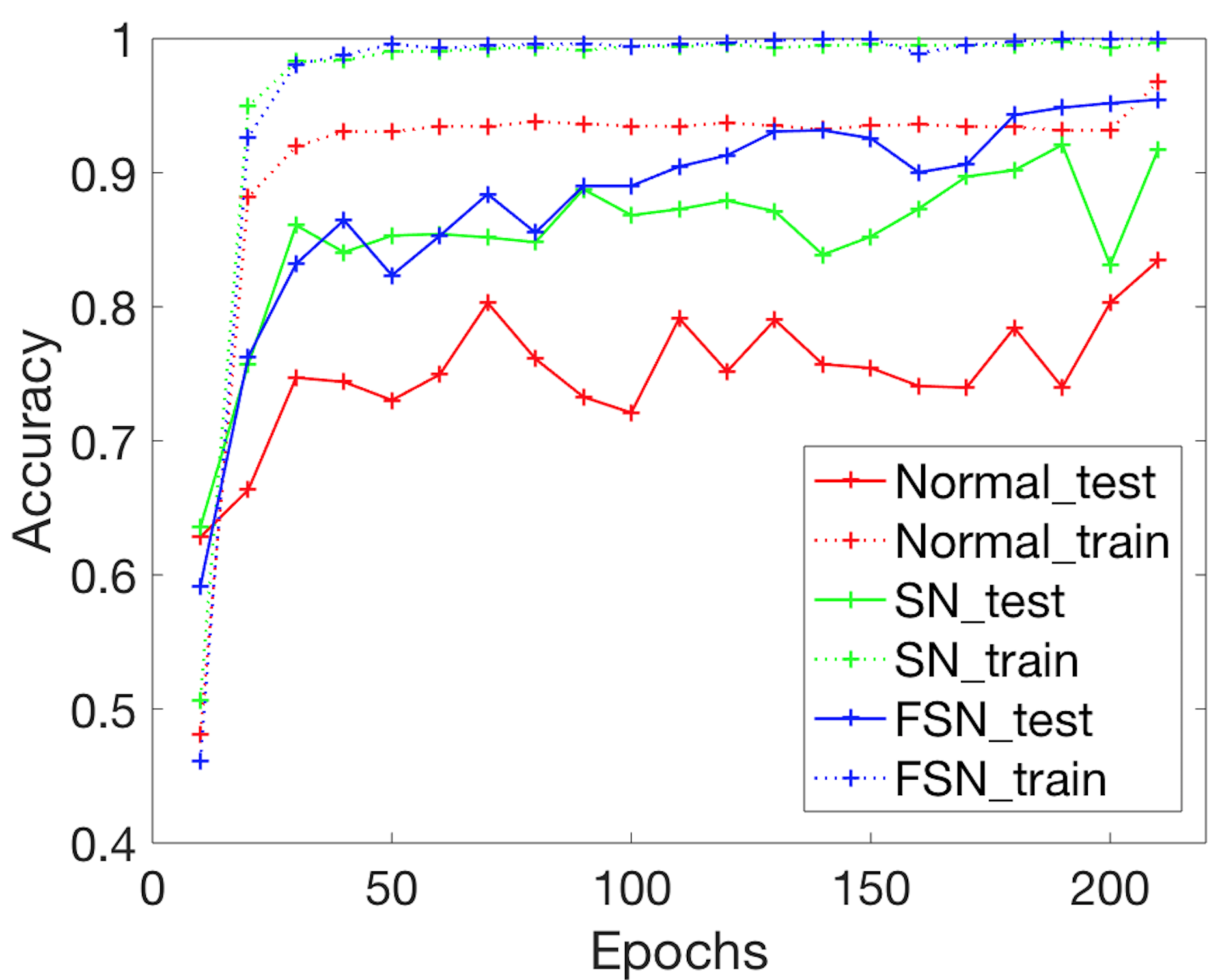}
\vspace{-0.05in}
\caption{Accuracy curve of VGG16 on GTSB}
\label{fig:label6}
\end{figure}

Table~\ref{Tb1} indicates the basic functionality performance of three training methods. We also plot the training curve for accuracy updates in Figure~\ref{fig:label6}. In general, all three methods possess good training accuracy. The Normal approach achieved the fastest speed but when it comes to test accuracy, it lagged behind the other two due to spectral normalization's ability to enhance model's generalizability~\cite{SNCNN}. SN promote the test accuracy by $8.27\%$ but also yields the highest time cost. FSN further improves the test accuracy by $3.66\%$ and reduce average training time by $60.5\%$ compared with to state-of-the-art (SN).

\vspace{-0.1in}
\vspace{-0.1in}

Table~\ref{Tb2} shows the performance of three models against bounded attacks. We choose four different attack algorithms: FGSM~\cite{FSGM}, CW-$L_2$~\cite{Carlini}, JSMA~\cite{JSMA} and DeepFool~\cite{DeepFool}. For consistency $\lambda = 0.01$ for both SN and FSN here. The hyperparameters for attacks are provided in the table.

%\vspace{-0.05in}
\begin{table}[tp]
    \caption{Bounded Attack on GTSB}
    %\vspace{20pt}
    \centering
    \begin{tabular}{c|ccccc}
        \hline
        \thead[l]{Methods} & \thead[l]{N/A}& \thead[l]{FGSM \\($\varepsilon$ = .1)} & \thead[l]{CW-$L_2$\\(c = .01)} & \thead[l]{JSMA \\$(\theta,\gamma = .1,1)$} & \thead[l]{DeepFool\\ $\varepsilon=1e-6$}\\
        \hline
        Normal   & 83.45 & 61.04 & 45.62 & 77.58 & 40.06 \\
        SN   & 91.72 & 88.25 & 76.31 & 88.23 & 69.30\\
        FSN & 95.46 & 92.92 & 87.05 & 91.57 & 88.94 \\
        \hline       
    \end{tabular}
    \label{Tb2}
    \vspace{-0.15in}
\end{table}

\begin{figure}[htbp]
\centering
\subfigure[Performance of three models towards unbounded FGSM attack] {\includegraphics[height=1.3in,width=1.7in]{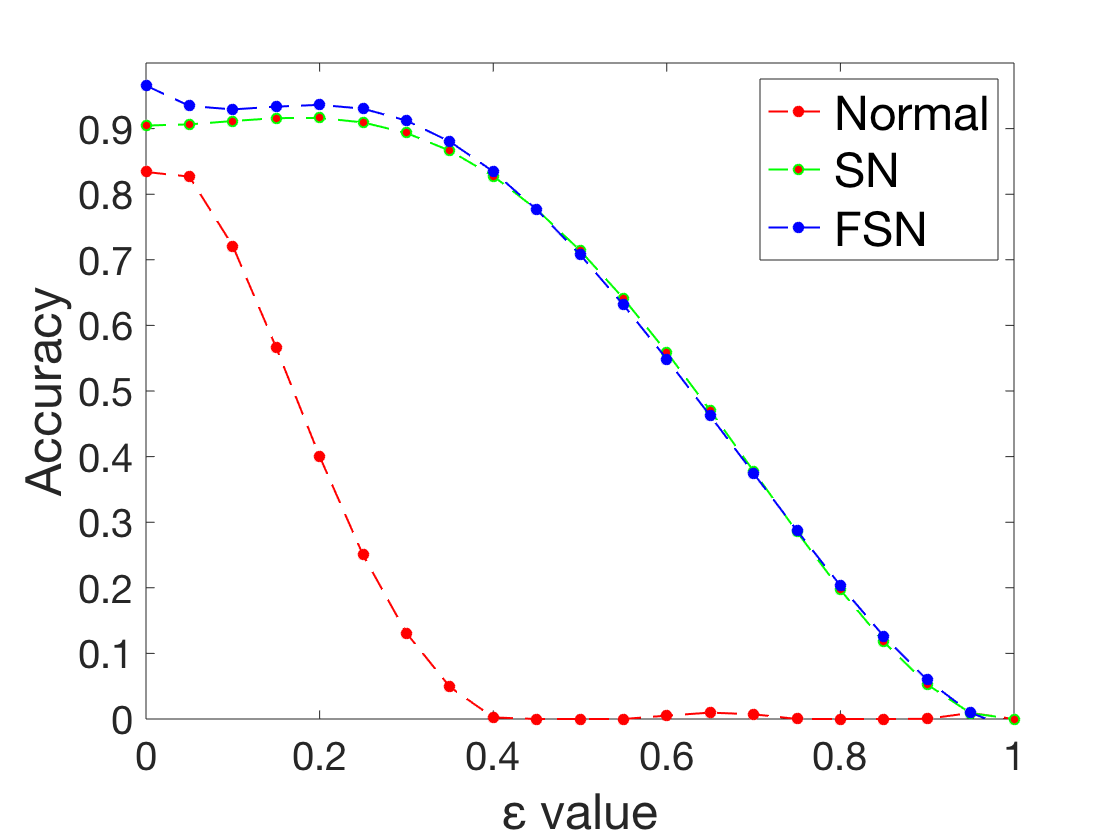}}
\subfigure[FSN's performance with different $\lambda$ and attack algorithm] {\includegraphics[height=1.3in,width=1.7in]{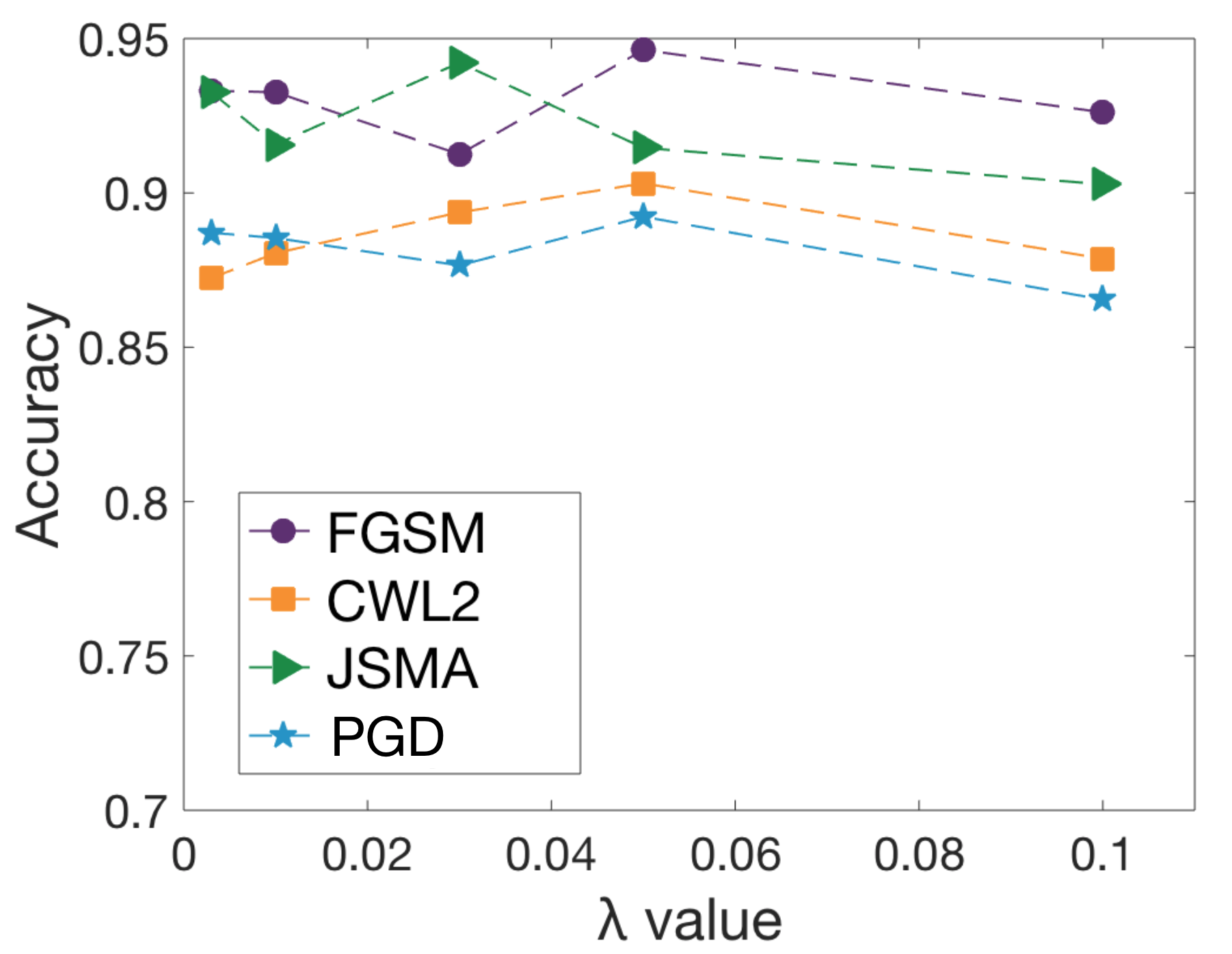}}
\vspace{-0.1in}
\caption{Performance with varying hyperparameter (GTSB)}
\label{fig:label7}
\end{figure}

As we can see, our proposed method (FSN) provides the best robustness. The Normal and SN method appeared fragile in the face of powerful attacks like CW-$L_2$ and DeepFool, while FSN still retained an acceptable accuracy. For lightweight attacks, especially gradient-based ones (like FGSM), FSN is almost unaffected. For unbounded attack, we applied unbounded incremental $\varepsilon$ value in FGSM from 0 to 1. It starts to break detection (accuracy $<$ 50\%) as presented in Figure~\ref{fig:label7}(a), but the bisection parameter value of FSN($\varepsilon= .64$) has to be nearly three times large of that for Normal method($\varepsilon= .21$). Finally, FSN's broad robust performance across a wide range of hyperparameter $\lambda$ is demonstrated in Figure~\ref{fig:label7}(b) where we plot accuracy under all attacks with $\lambda$ varying from 0.01 to 0.1. The average accuracy here is $91.46\%$ with standard deviation of 0.0729. In the worst case, model trained with FSN still performed well with accuracy above $85\%$.

\vspace{-0.2in}
\vspace{-0.1in}

\subsection{Case Study: Udacity Self-Driving (USD)}
We trained Lenet-5 on USD's steering angle image dataset which has more than 5000 autonomous vehicle's real-time screenshot along with labels indicating the corresponding ``Left", ``Right" or ``Center" prediction for next wheel steering operation. Compared with VGG16, Lenet-5 is a lightweight network with only five convolution layers and one fully connected layer. The training process consists of 500 epochs with a mini-batch size of 128. For every 50 epochs, we randomly shuffle sample and split $80\%$ as training set and $20\%$ as test.

\begin{table}[htbp]
    \caption{Training time and test accuracy of Lenet-5}
    %\vspace{20pt}
    \centering
    \begin{tabular}{c|cc}
        \hline
        \thead[l]{Methods} & \thead[l]{Time(s/epoch)}& \thead[l]{Best Test Accuracy(\%)} \\
        \hline
        Normal  & 0.13369   & 99.71 \\
        SN      & 0.25870   & 98.57 \\
        FSN     & 0.18284   & 99.43 \\
        \hline       
    \end{tabular}
    \label{Tb3}
\end{table}

\begin{figure}[h]
\vspace{-0.2in}
\centering
\includegraphics[scale =0.18]{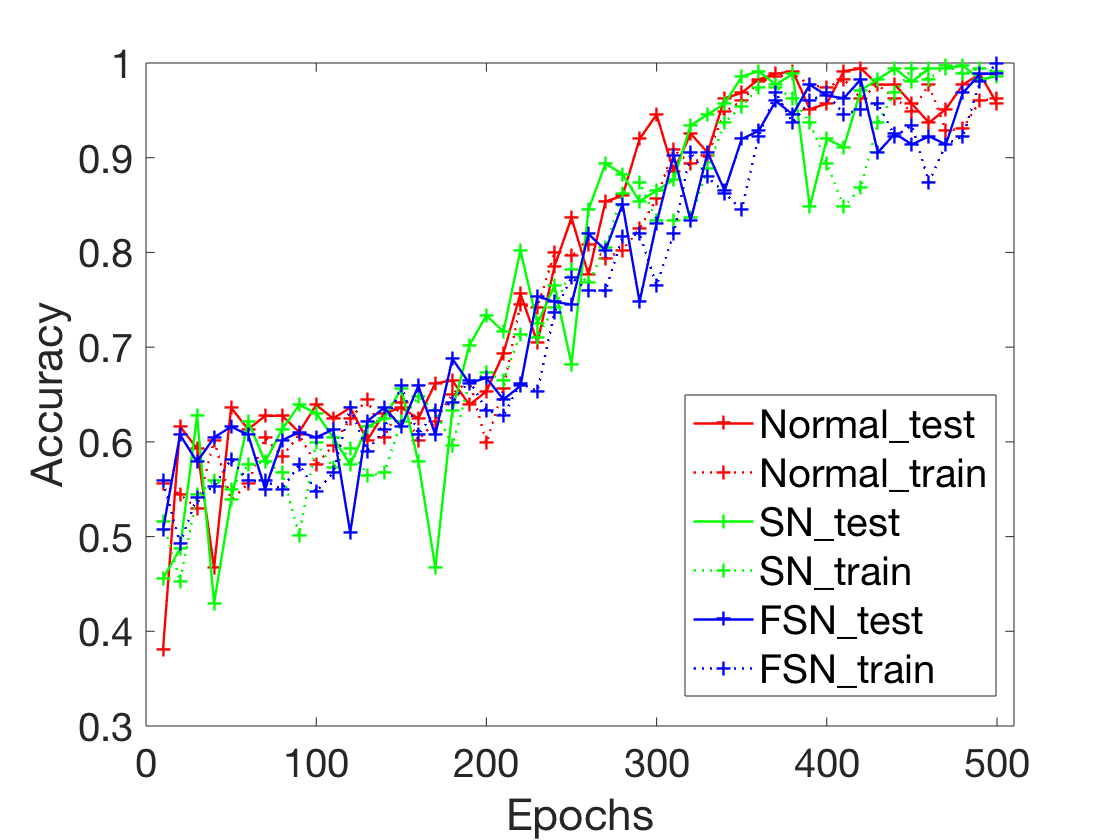}
\vspace{-0.1in}
\caption{Accuracy curve of Lenet-5 on USD}
\label{fig:label8} 
\vspace{-0.1in}
\end{figure}

Table~\ref{Tb3} and Figure~\ref{fig:label8} present performance results. Again, Normal method provides the fastest training speed, while FSN is nearly $41.4\%$ faster than SN. Note that the improvement of speed here is not as high as we got from GTSB case, and we consider this difference of acceleration amplitude is caused by more convolution layers involved in VGG16 than Lenet-5. For accuracy, due to Lenet-5's excellent capability in image recognition, all three models achieved high level of accuracy.

\begin{figure}[htbp]
\centering
\subfigure[Performance of three models under unbounded FGSM attack] {\includegraphics[height=1.3in,width=1.7in]{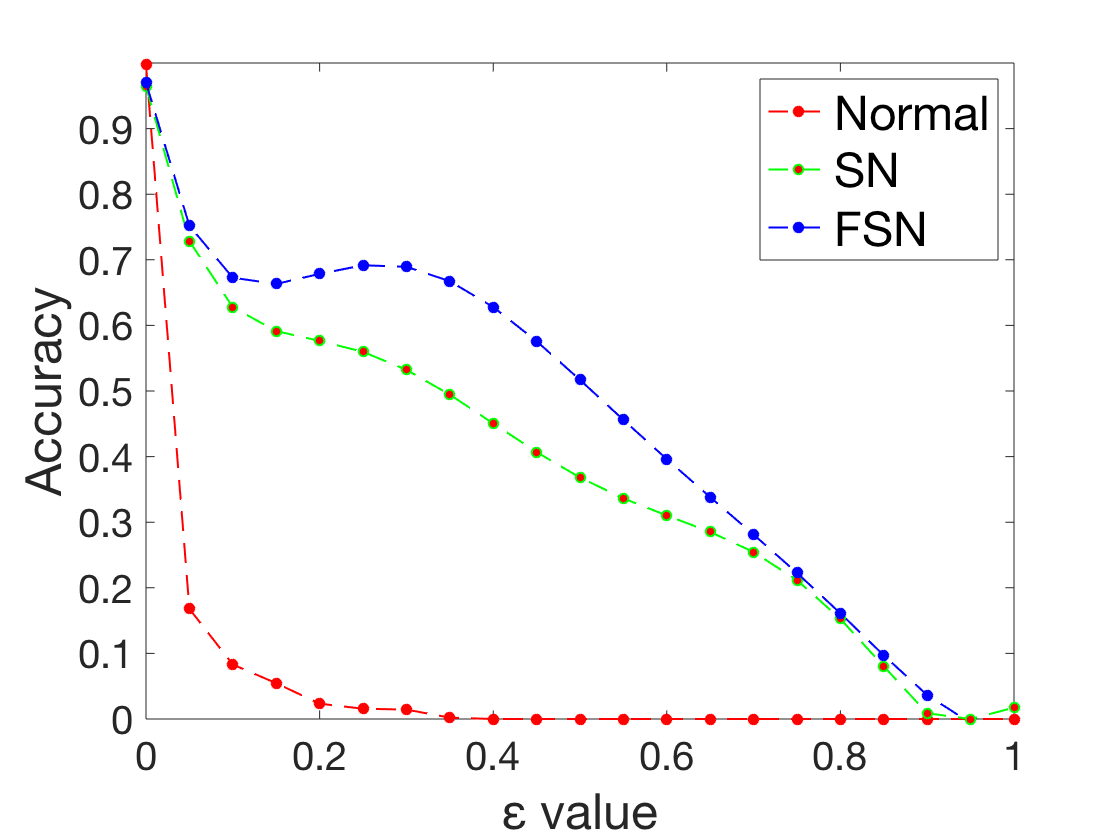}}
\subfigure[FSN's performance with different $\lambda$ and attack algorithm] {\includegraphics[height=1.3in,width=1.7in]{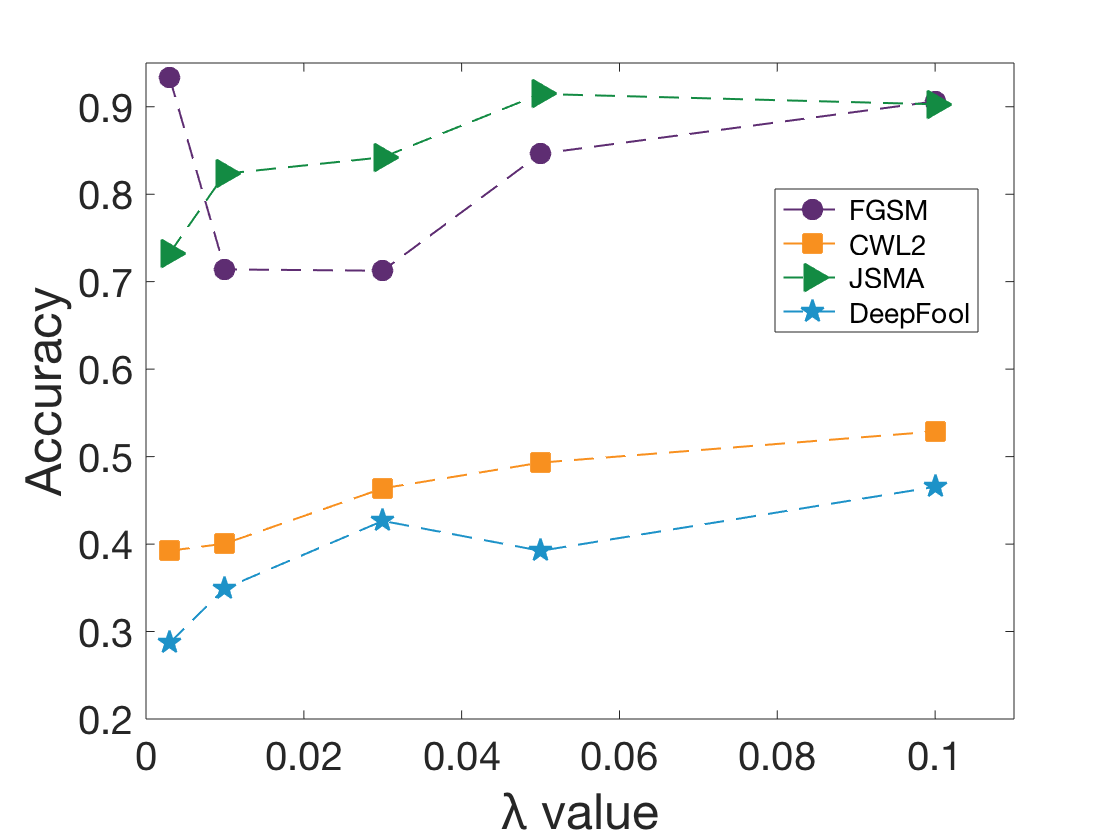}}
\vspace{-0.1in}
\caption{Performance with varying hyperparameters (USD)}
\vspace{-0.1in}
\label{fig:label9}
\end{figure}

 The robustness is evaluated against four attack algorithms. As expected, Normal method behaved extremely vulnerable. Even in $\varepsilon = 0.1$ case, a $8.3\%$ accuracy against FGSM attack was obtained. FSN refined it to $71.4\%$ which gives an 61.1\% average improvement on accuracy. When it comes to unbounded attack, the Normal method's outcome immediately drops below $50\%$ when $\varepsilon = 0.1$, while SN and FSN extended it to $0.4$ and $0.5$, respectively. Figure~\ref{fig:label9}(b) shows the accuracy after applying various regularization factor of $\lambda$. The average accuracy is 62.63\% with standard deviation of 0.2252. If we consider only CW-$L_2$ and DeepFool, the average drops to $41.99\%$ with standard deviation of $0.0713$.  
 
 \begin{table}[H]
    \caption{Bounded Attack on USD}
    %\vspace{20pt}
    \centering
    \begin{tabular}{c|ccccc}
        \hline
        \thead[l]{Methods} & \thead[l]{N/A}& \thead[l]{FGSM \\($\varepsilon$ = .1)} & \thead[l]{CW-$L_2$\\(c = .01)} & \thead[l]{JSMA \\$(\theta,\gamma = .1,1)$} & \thead[l]{DeepFool\\ $\varepsilon=1e-6$}\\
        \hline
        Normal   & 99.71 & 8.3 & 5.62 & 17.58 & 4.06 \\
        SN       & 98.57 & 69.9 & 34.34 & 78.23 & 29.30\\
        FSN      & 99.43 & 71.4 & 40.05 & 82.34 & 34.94 \\
        \hline       
    \end{tabular}
    \label{Tb4}
\end{table}

%The improvement of given model's robustness against CW-$L_2$ and DeepFool is not as good as the improvement in GTSB. This is because Lenet-5 possess relatively fewer layers, and therefore, the summation of spectral norm penalty is smaller than that in GTSB. Fortunately, typical AD systems are significantly more complex than Lenet-5, where our approach is able to achieve significant improvement in robustness. 
 
\vspace{-0.1in}

%% file: sections/conclusion.tex
\section{Conclusion}
% In this paper, we investigate into adversarial attack scenarios and introduce an approximate regularization strategy, the fast spectral norm regularization. It utilized the spatial separation of traditional convolution layers to drastically reduce time complexity in computation. And by substituting power method with Fourier method, an up to $60\%$ faster spectral norm computation process helped us to obtained spectral norm. By appending the approximately computed spectra norm to loss function as a penalty term, it obviously reduced given model's sensitivity against input perturbation. 

% As demonstrated in experiment results, models trained with our method showed great improvement in robustness against various adversarial attacks. 

% For example, an attacker has to conceive perturbation's scale in a ratio of $2.57$ larger to cheat our enhanced model, which means the difference between adversarial samples and begin samples is enlarged and it becomes easier to be detected beforehand. 

% Our proposed regularized training approach is expected to play a crucial role in improving the robustness of future systems against adversarial attacks.

Adversarial attacks are vital threats towards deep neural networks.  In this paper, we investigated adversarial attack scenarios and proposed a novel regularized training strategy, the fast spectral normalization (FSN). Such a neural network regularization technique made three important contributions. (1) FSN utilized the spatial separation of convolution layers as well as Fourier transform to drastically (up to $60\%$) reduce training time compared with spectral normalization. (2) FSN is algorithm-agnostic, easy to implement, and applicable across a wide variety of datasets. (3) Experimental evaluation using three popular benchmarks demonstrated that models trained with FSN provide significant improvement in robustness for bounded, unbounded as well as transferred threats, which can protect applications from various adversarial attacks. Compared with the state-of-the-art, our approach ensures that the model works correctly unless the adversarial samples are drastically different from the original samples.
%and the loss of accuracy for spectral normalization is degraded by and the replacement of power iteration. 
% We observed a promising compatibility between our method and Residual Network while investigating the instability of the CIFAR-10 dataset - as the residual units in Residual Network ensure the sparsity of convolution layers, it accelerates the process of layer separation and Alnordi iteration converges at the same time. Our method also provides the effect of weight regularization to alleviate overfitting problem, which is demonstrated using the GTSB dataset. 
Our proposed regularized training approach is expected to play a crucial role in improving the robustness of future systems against adversarial attacks.